\documentclass[10pt,twocolumn,letterpaper]{article}

\usepackage{wacv}
\usepackage{times}
\usepackage{epsfig}
\usepackage{graphicx}
\usepackage{amsmath}
\usepackage{amssymb}

\usepackage{booktabs}
\usepackage{dsfont}
\def\1{\mathds{1}}
\usepackage{enumitem}

\usepackage{authblk}

\usepackage[pagebackref=true,breaklinks=true,letterpaper=true,colorlinks,bookmarks=false]{hyperref}

\wacvfinalcopy 

\def\wacvPaperID{670} 

\ifwacvfinal\fi
\begin{document}

\title{Anchor Box Optimization for Object Detection}

\author[1]{Yuanyi Zhong\thanks{This work was partly performed when the first author was interning at Microsoft.}}
\author[2]{Jianfeng Wang}
\author[1]{Jian Peng}
\author[2]{Lei Zhang}
\affil[ ]{$^1$University of Illinois at Urbana-Champaign, $^2$Microsoft}
\affil[ ]{ \tt \small $^1$\{yuanyiz2, jianpeng\}@illinois.edu, $^2$\{jianfw, leizhang\}@microsoft.com }

\maketitle
\ifwacvfinal\thispagestyle{empty}\fi

\begin{abstract}
In this paper, we propose a general approach to optimize anchor boxes for object detection. Nowadays, anchor boxes are widely adopted in state-of-the-art detection frameworks. However, these frameworks usually pre-define anchor box shapes in heuristic ways and fix the sizes during training. To improve the accuracy and reduce the effort of designing anchor boxes, we propose to dynamically learn the anchor shapes, which allows the anchors to automatically adapt to the data distribution and the network learning capability. The learning approach can be easily implemented with stochastic gradient descent and can be plugged into any anchor box-based detection framework. The extra training cost is almost negligible and it has no impact on the inference time or memory cost. 
Exhaustive experiments demonstrate that the proposed anchor optimization method consistently achieves significant improvement ($\ge1\%$ mAP absolute gain) over the baseline methods on several benchmark datasets including Pascal VOC 07+12, MS COCO and Brainwash. Meanwhile, the robustness is also verified towards different anchor initialization methods and the number of anchor shapes, which greatly simplifies the problem of anchor box design. 
\end{abstract}

\section{Introduction}

Object detection plays an important role in many real world applications and recent years have seen great improvement in speed and accuracy thanks to deep neural networks \cite{Ren2015FasterRT, Redmon2017YOLO9000BF, Redmon2018YOLOv3AI, Liu2016SSDSS, Lin2018FocalLF}. 
Many modern deep learning based detectors make use of the anchor boxes (or default boxes), which serve as the initial guesses of the bounding boxes. 
These anchor boxes are densely distributed across the image, typically centered at each cell of the final feature map. 
Then the neural network is trained to predict the translation offsets relative to the cell center (sometimes normalized by the anchor size) and the width/height offsets relative to the anchor box shape, as well as the classification confidence. 

One of the critical factors here is the design of the anchor width and the anchor height, and most of prior approaches determine these values by ad-hoc heuristic methods. 
For instance, in Faster R-CNN\cite{Ren2015FasterRT}, the anchor shapes are hand-chosen to have 3 scales ($128^2$, $256^2$, $512^2$) and 3 aspect ratios ($1\mathbin{:} 1$, $1\mathbin{:}2$, $2\mathbin{:}1$). 
When applying the general object detectors on specific domains, the anchor shapes have to be manually tweaked to improve accuracy. For text detection in~\cite{LiaoSB18}, the aspect ratios also include $5\mathbin{:}1$ and $1\mathbin{:}5$, since texts could exhibit wider or higher ranges than generic objects. 

Once the anchor shapes are determined, the sizes will be fixed during training. This might be sub-optimal since it disregards the augmented data distribution in training, the characteristics of the neural network structure and the task (loss function) itself. Improper design of the anchor shapes could lead to inferior performance in specific domains. 

To address the issue, we propose a novel anchor optimization approach that can automatically \emph{learn} the anchor shapes during training. This could leave the choice of anchor shapes completely in the hands of learning, such that the learned shapes can adapt better to the dataset, the network and the task without much human interference. The learning approach can be easily implemented with stochastic gradient descent (SGD) and could be plugged into any anchor box based detection framework. 
To verify the idea, we conduct extensive experiments on several benchmark datasets including Pascal VOC 07+12, MS COCO and Brainwash, building upon both YOLOv2 and Faster-RCNN frameworks. 
The results strongly demonstrate that optimizing anchor boxes can significantly improve the accuracy ($\ge 1\%$ mAP absolute gain with YOLOv2) over the baseline method.
Meanwhile, the robustness is also verified towards different anchor box initializations and the improvement is consistent across different number of anchor shapes, which greatly simplifies the problem of anchor box design.

The main contributions are summarized as follows:
\begin{itemize}[itemsep=-1pt]
    \item We present a novel approach to optimize the anchor shapes during training, which, to the best of our knowledge, is the first time to treat anchor shapes as trainable variables without modifying the inference network. 
    \item We demonstrate through extensive experiments that the anchor optimization method not only learns the appropriate anchor shapes but also boosts the detection accuracy of existing detectors significantly. 
    \item We also verify that our method is robust towards initialization, so the burden of handcrafting good anchor shapes for specific dataset is greatly lightened.
\end{itemize}

\section{Related Work}\label{sec:related}
The modern object detectors usually contain two heads: one for classification and the other for localization. The classification part is to predict the class confidence, while the localization part is to predict the bounding box coordinates. Based on how the location is predicted, we roughly categorize the related work into two genres: relative offset prediction based on pre-defined anchor boxes \cite{Szegedy2014ScalableHO,Liu2016SSDSS}, and absolute offset prediction \cite{Redmon2016YouOL,TychsenSmith2017DeNetSR,Law2018CornerNetDO}.

\subsection{Relative Offset Prediction}
The network predicts the offset relative to the pre-defined anchor boxes, which are also named as {\it default boxes} \cite{Liu2016SSDSS}, and {\it priors} \cite{Szegedy2014ScalableHO}. 
These boxes serve as the initial guess of the bounding box positions. The anchor shapes are fixed during training and the neural network learns to regress the relative offsets. 
Assume $(\Delta^{(x)},\Delta^{(y)},\Delta^{(w)},\Delta^{(h)})$ are the neural net outputs, one typical approach \cite{Ren2015FasterRT, Liu2016SSDSS} is to express the predicted box as 
$(a^{(x)} + \Delta^{(x)} a^{(w)}, a^{(y)} + \Delta^{(y)} a^{(h)}), a^{(w)}e^{\Delta^{(w)}}, a^{(h)}e^{\Delta^{(h)}})$ 
where $a^{(w)}, a^{(h)}$ are the pre-defined anchor width and height, $a^{(x)}, a^{(y)}$ are the anchor box center. The former two numbers represent the box center and the latter two represent the box width and height. 
Thus, one critical problem is to design the anchor shapes. 

In Faster R-CNN \cite{Ren2015FasterRT}, the anchor shapes are chosen with 3 scales ($128^2$, $256^2$, $512^2$) and 3 aspect ratios ($1\mathbin{:}1$, $1\mathbin{:}2$, $2\mathbin{:}1$), yielding 9 different anchors at each output sliding window position. 
In Single Shot MultiBox detector (SSD) \cite{Liu2016SSDSS}, the anchor boxes also have several scales on different feature map levels and aspect ratios including $1\mathbin{:}3$, $3\mathbin{:}1$ in addition to $1\mathbin{:} 1$, $1\mathbin{:}2$, $2\mathbin{:}1$. In YOLOv2~\cite{Redmon2017YOLO9000BF}, the anchor shapes are not handcrafted, but are the $k$-Means centroids with IoU as the similarity metric. 
With the localization regression result on top of the anchors, extra regressors can be used to further refine the candidate bounding boxes, possibly through multiple stages, e.g. in Cascade RCNN\cite{CaiV18}, RefineDet\cite{zhang2018single}, and Guided Anchoring\cite{wang2019region}.

When the general object detection framework is applied to specific problems, the anchor sizes have to be hand-tuned accordingly. For example, text detection in~\cite{LiaoSB18} employs the aspect ratios $5\mathbin{:}1$ and $1\mathbin{:}5$ as well as $1\mathbin{:}1$, $1\mathbin{:}2$, $2\mathbin{:}1$, $1\mathbin{:}3$, $3\mathbin{:}1$, since texts could exhibit wider or higher range than generic objects. 
For face detection in~\cite{NajibiSCD17,ZhangZLSWL17}, the aspect ratios only include $1\mathbin{:}1$ since face is roughly in a square shape. 
For pedestrian detection in \cite{zhang2016faster}, a ratio of 0.41 based on \cite{Dollr2012PedestrianDA} is adopted for the anchor boxes.
As suggested in \cite{zhang2016faster}, inappropriate anchor boxes could 
degrade the accuracy. 

To ease the effort of anchor shape design, the most relevant work might be
MetaAnchor\cite{Yang2018MetaAnchorLT}. Leveraging neural network weight prediction, the anchors are modeled as functions parameterized by an extra neural network and computed from customized prior boxes. The mechanism is shown to be robust to anchor settings and bounding box distributions. 
However, the method involves an extra network to predict the weights of another network, resulting in extra training effort and inference time, and also needs to choose a set of customized prior boxes by hand. Comparatively, our method can be easily embedded into any detection framework without extra network, and has negligible impact on the training time/space cost and no impact on the inference time/space. 

\subsection{Absolute Offset Prediction}
Another research effort is to directly predict the absolute location values rather than its position and size relative to pre-defined anchor boxes. The YOLO~\cite{Redmon2016YouOL} belongs to this spectrum but was improved by YOLOv2~\cite{Redmon2017YOLO9000BF} with anchor-based approach. In DeNet \cite{TychsenSmith2017DeNetSR}, the network outputs the confidence of each neuron belonging to one of the bounding box corners, and then collects the candidate boxes by Directed Sparse Sampling. More recently, CornerNet~\cite{Law2018CornerNetDO} proposed detecting objects by the top-left and bottom-right keypoint pairs, and introduces the corner pooling operation to better localize corners. While these anchor-free methods form a promising future research direction, yet anchor-based methods still seem to achieve the mainstream best accuracy in the public benchmarks.
 
\begin{figure*}[t]
\begin{center}
\includegraphics[width=0.95\linewidth]{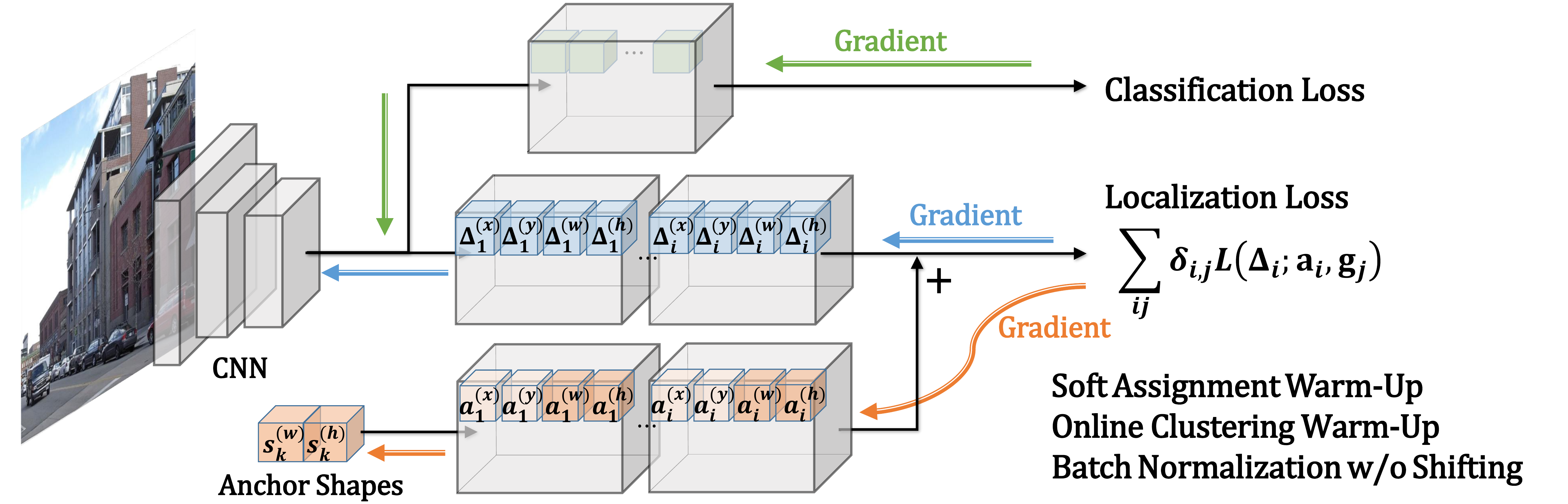}
\end{center}
\caption{An illustration of the anchor optimization process. The localization loss is to minimize the error between the ground-truth box and the predicted offset relative to the anchor box (with the batch norm trick). The error is back-propagated to the anchor shapes and the CNN parameters to automatically learn the anchor size. The anchor shape is warmed up by the soft assignment and the online clustering.}
\label{fig_arch}
\end{figure*}

\section{Proposed Approach}\label{sec:approach}

We first present an overview of existing anchor-based object detection frameworks, and then describe the proposed anchor optimization approach.

\subsection{Object Detection Overview}

In state-of-the-art object detection frameworks, the training 
is usually done by minimizing a combination of bounding box localization loss and category classification loss. 

\subsubsection{Localization Loss}
For one feature map with $A$ different anchor shapes from the network, each spatial cell location could correspond to $A$ anchor boxes centered at that cell. Thus the total number of anchor boxes is $N\triangleq A\times H_f\times W_f$, where $H_f$ and $W_f$ are the feature map height and width. 
Stacking all the anchor boxes, we can denote by $\mathbf{a}_i = (a_i^{(x)}, a_i^{(y)}, a_i^{(w)}, a_i^{(h)})$ the $i$-th ($i \in \{1, \cdots, N\}$) anchor box, where $a_i^{(x)}$ and $a_i^{(y)}$ represent the center of the box and $a_i^{(w)}$ and $a_i^{(h)}$ represent the width and height respectively. 
For multiple feature maps as in \cite{Lin2018FocalLF, Liu2016SSDSS}, similar notations can also represent all the anchor boxes stacked together. 
Note since we only have $A$ different anchor shapes, the value of $a_i^{(w)}$ and $a_i^{(h)}$ only have $A$ instead of $N$ different values. 
The anchor center $a_i^{(x)}, a_i^{(y)}$ are usually linearly related to the spatial location on the feature map. The shape $a_i^{(w)}, a_i^{(h)}$
are pre-defined and remain constant during training in existing work. 

Let $\mathbf{\Delta}_i = (\Delta_i^{(x)}, \Delta_i^{(y)}, \Delta_i^{(w)}, \Delta_i^{(h)})$ be
the network output for the $i$-th anchor box. 
Then, the localization loss is to align the network prediction to the ground-truth bounding box coordinates with respect to the anchor box. Specifically, the loss for the $i$-th anchor box could be written as 
\begin{align}
    L_{loc} = \delta_{i, j}L(\mathbf{\Delta}_i; \mathbf{a}_i, \mathbf{g}_{j}),
\end{align}
where $\mathbf{g}_j=(g_j^{(x)}, g_j^{(y)}, g_j^{(w)}, g_j^{(h)})$ are the $j$-th ground-truth box and $\delta_{i, j}$
measures how much the $i$-th anchor should be responsible to the $j$-th ground-truth.

The value of $\delta_{i,j}$ is usually restricted in $\{0,1\}$, where 1 indicates that the $i$-th anchor box is responsible for the $j$-th ground-truth box. For example in \cite{Ren2015FasterRT, Lin2018FocalLF, Liu2016SSDSS}, $\delta_{i, j}$ is $1$ if the IoU between the anchor box and the ground-truth is larger than a threshold e.g. 0.5 or the anchor box is the one with the largest overlap with the ground-truth. In YOLOv2~\cite{Redmon2017YOLO9000BF}, $\delta_{i, j}$ is $1$ if the anchor box is located in the same spatial location as and has the largest IoU with the ground-truth.

The form of the localization loss could be $L^2$~\cite{Redmon2017YOLO9000BF}, or smoothed $L^1$ (i.e. Huber loss)~\cite{Ren2015FasterRT, Liu2016SSDSS}.
For example, the L2 loss $L(\mathbf{\Delta}_i; \mathbf{a}_i, \mathbf{g}_j)$ is the sum of $L_{i, j}^{(x, y)}$ and 
$L_{i, j}^{(w, h)}$ with 

{\small
\begin{align}
L_{i, j}^{(x, y)} = & (\Delta_i^{(x)} + a_{i}^{(x)} - g_{j}^{(x)})^2 + 
    (\Delta_i^{(y)} + a_{i}^{(y)} - g_{j}^{(y)})^2 
    \label{eq_loss_xy}\\
L_{i, j}^{(w, h)} = & (\Delta_i^{(w)} + \hat{a}_i^{(w)}  - \hat{g}_j^{(w)})^2 +     (\Delta_i^{(h)} + \hat{a}_i^{(h)} - \hat{g}_j^{(h)})^2
    \label{eq_loss_wh}\\
\hat{a}_i^{(w)} \triangleq & \log(a_i^{(w)}), \text{~~~} \hat{a}_i^{(h)} \triangleq  \log(a_i^{(h)}) \\
\hat{g}_{j}^{(w)} \triangleq & \log(g_{j}^{(w)}), \text{~~~} 
\hat{g}_j^{(h)} \triangleq \log(g_{j}^{(h)}) .
\end{align}}

The width and height follow the conventional log encoding scheme. 
They appear explicitly in the wh-loss Eqn.~\ref{eq_loss_wh}. This enables direct gradient computation on $a^{(w)}_j, a^{(h)}_j$, which is the key of our anchor optimization method.

\subsubsection{Classification Loss}
For each anchor box, the network also outputs the confidence score to identify which class it belongs to. 
In training, usually cross entropy loss is employed, e.g. in \cite{Ren2015FasterRT, Liu2016SSDSS, Redmon2016YouOL, Redmon2017YOLO9000BF}. One improved version is the focal loss \cite{Lin2018FocalLF}, which tackles the class imbalance issue. 

To handle the background, one can add an extra background class in the cross entropy loss, e.g. in \cite{Liu2016SSDSS, Ren2015FasterRT}. Another approach is to learn a class-agnostic objectness score to identify if there is an object, e.g. in YOLOv2\cite{Redmon2017YOLO9000BF} and the RPN of Faster R-CNN\cite{Ren2015FasterRT}.

\subsection{Anchor Box Optimization}\label{sec:opt}
By combining the localization loss and the classification loss in $L$, we can write the optimization problem as 
\begin{align}
    \min_{\theta} L(\theta) \label{eq_min1}
\end{align}
where $\theta$ is the neural network parameters. In existing methods, the anchor shapes are treated as constants. For all the $N$ anchor boxes $\mathbf{a}_i$, we extract all the distinct anchor shapes and denote them by
$(s_k^{(w)}, s_k^{(h)})_{k=1}^{A}$. There are $A$ of them, for example, $A=5$ by default in YOLOv2.
In the proposed method, we treat them as trainable variables of the optimization problem Eqn.~\ref{eq_min2}. 
\begin{align}
    \min_{\theta, \{ (s_k^{(w)}, s_k^{(h)}) \}_{k=1}^{A} } L \left( \theta, \{ (s_k^{(w)}, s_k^{(h)}) \}_{k=1}^{A} \right)
        \label{eq_min2}
\end{align}
Obviously, Eqn.~\ref{eq_min2} guarantees a smaller optimal loss value than Eqn.~\ref{eq_min1} since the trainable variable set is enlarged (so is the feasible solution set).   
The anchor shape values can be adjusted towards the goal of lowering the overall loss value. Moreover, with the learned optimal anchor shapes, the magnitudes of the offsets (residuals)  $\mathbf{\Delta}_i$ can become smaller, which might make the regression problem easier.

The key idea is summarized in Fig.~\ref{fig_arch}. 
Following common practice, we use the back-propagation to solve the optimization problem.
Instead of learning $s_k^{w}$ and $s_k^{h}$, we learn 
$\hat{s}_k^w \triangleq  \log(s_k^w)$ and 
$\hat{s}_k^h \triangleq  \log(s_k^h)$ because of equivalence and simplicity. 
For a single training image, the derivative of the loss function with respect to $\hat{s}_k^{(w)}$ can be computed as
\begin{align}
    \frac{\partial L}{\partial \hat{s}_k^{(w)}} &\propto  \sum_{i, j} \delta_{i, j} \left( \frac{\partial}{\partial \hat{s}_k^{(w)}} L^{(w, h)}_{i, j} \right) \\
    &\propto \sum_{i, j} \delta_{i, j} \left(\Delta_{i} + \hat{a}_i^{w} - \hat{g}_j^w \right)\delta(\hat{a}_i^w = \hat{s}_k^w), \label{eq_da}
\end{align}
where $\delta(\hat{a}_i^w = \hat{s}_k^w)$ is 1 if $\hat{a}_i^w$ corresponds
to $\hat{a}_j^w$, and is 0 otherwise.
Similarly, we can have the derivative with respect to the anchor height $\hat{a}_k^{(h)}$.

In one training iteration of the mini-batch SGD, we firstly assign the ground-truth boxes to the anchors, i.e. computes $\delta_{i, j}$ for each image. Then, with $\delta_{i, j}$ fixed, back-propagate the error signal to all remaining parameters including the anchor shapes $\hat{s}_k^w, \hat{s}_k^h$.
To calculate the value of $\delta_{i, j}$, we use the IoU as the metric following convention~\cite{Ren2015FasterRT, Liu2016SSDSS, Redmon2017YOLO9000BF}. If we otherwise use $L^2$ distance in the log space of width and height as metric to compute $\delta_{i, j}$, the metric aligns more closely with the loss. However, we empirically find that $L^2$ distance and IoU end up with similar performances and similar anchor shapes.

Finally, we emphasize that the proposed approach is different from learnable bias terms in bounding box prediction. It is analogous to the difference between the k-means algorithm (which iterates between computing centers and reassigning instances) and only computing centers. With our approach, the ground truth box to anchor assignments will gradually change as the anchor shapes are being optimized, to lower the overall cost; While with learnable bias terms, the assignments are fixed throughout training.

To further facilitate automatic learning of anchor shapes, we introduce the following three training techniques.

\subsubsection{Soft Assignment Warm-Up}
In some extreme situation, the ground-truth boxes could be all very small or large, as a result, only one anchor shape is activated and the remaining ones are never used during training. To address this issue, we propose to adopt a soft assignment approach at the \emph{early training stage}. That is
\begin{align}
\delta_{i, j} = \text{softmax} ( -\text{dist}( \mathbf{a}_i, \mathbf{g}_j ) / T), \end{align}
where the $\text{softmax}$ is over all anchor boxes at the same spatial cell. The temperature $T$ is annealed from 2 to 0 in the first few training steps (1500 iterations in our experiments). With non-zero assignment values, all anchor shapes can join into the learning procedure. 
After the warm-up, it falls back to the original assignment scheme.

\subsubsection{Online Clustering Warm-Up}
Motivated by the $k$-means approach in~\cite{Redmon2017YOLO9000BF}, we augment the loss function
with an extra online clustering term during again the \emph{early stage} of training. 
This term minimizes the squared $L^2$ distance between the anchor shapes and the ground-truth box shapes and can be written as
\begin{align}
L_{\text{aug}} & = L + \lambda   \frac{1}{2\mathcal{N}} \sum_{i, j} \delta_{i, j} T_{i, j}, \label{eq_aug}  \\
T_{i, j} & \triangleq (\hat{a}^{(w)}_i - \hat{g}_j^{(w)})^2 + (\hat{a}^{(h)}_i - \hat{g}_j^{(h)})^2, \label{eqn:T} 
\end{align}
where $\mathcal{N}$ is $\sum_{i, j}\delta_{i, j}$ for normalization.

The coefficient $\lambda$ is linearly annealed from 1 to 0 during the early stage of training (first 1500 iterations in experiments) to warm up the learning of anchors. 
The underlying idea is that the $k$-means approach could serve as a good starting point. 
This makes the method more robust to initialization and faster to converge. 
In the early training stage, the clustering term could quickly tune the anchors to (near) k-means centroids. Then, the original loss of $L$ in Eqn.~\ref{eq_min2} begins to show more influence. The anchor shapes are thus able to adapt more closely to data distribution and network predictions, following gradients from the original loss. 

\subsubsection{Batch Normalization without Shifting} 
With the online clustering term in Eqn.~\ref{eqn:T}, the network output $\Delta_i$ tends to have a zero mean potentially following Gaussian distribution. 
To further reduce the learning difficulties, we apply the batch normalization \cite{Ioffe2015BatchNA} on the output of
$\Delta_{i}^{w}$ and $\Delta_i^{h}$ without the shifting parameters.
That is, the network output is normalized to zero mean and unit variance, followed by scaling operation but without the shift operation. 
This could enforce the zero mean distribution and make the training converge faster. 

\section{Experiments}\label{sec:exp}

We present the implementation details and the experiment results on the popular Pascal VOC 07+12 \cite{Everingham2009ThePV} and MS-COCO 2017 datasets \cite{Lin2014MicrosoftCC}, along with {\it{Brainwash}}\cite{Stewart2016EndtoEndPD}, a head detection dataset, to demonstrate the effectiveness of the proposed anchor optimization method.

\subsection{Implementation Details}

Since the proposed approach for optimizing anchors is quite general, it can be applied to most anchor-based object detection frameworks. 
We choose the YOLOv2~\cite{Redmon2017YOLO9000BF}, a classic one-stage detector, as the main test-bed to demonstrate the effectiveness. 
Extensions to other detectors should be straightforward, such as 
the RPN in Faster R-CNN~\cite{Ren2015FasterRT},
Feature Pyramid Network (FPN)~\cite{Lin2017FeaturePN}, SSD \cite{Liu2016SSDSS} and RetinaNet \cite{Lin2018FocalLF}. The applicability on the former two is further briefly explored.

The network consists of a DarkNet-19 backbone CNN pretrained on ImageNet classification, and several conv detection heads. With $A = 5$ anchor shapes, the last conv layer outputs a feature map of $ 5 \times (4 + 1 + C) $ channels, corresponding to 4 coordinate regression outputs ($\Delta_i$), 1 class-agnostic objectness score, and $C$ category scores for each anchor box. We also employ the same data augmentation techniques as in YOLOv2, including random jittering, scaling, and random hue, exposure, saturation change. The same loss weights are used to balance the localization, the objectness and the classification losses.

During testing, an image is resized to a specified size (e.g. 416-by-416) and fed into the detector. For each anchor box $\mathbf{a}_i$ and the corresponding output $\mathbf{\Delta}_{i}$, the output bounding box is $(a_i^{(x)} + \Delta_{i}^{(x)}, a_i^{(y)} + \Delta_{i}^{(y)}, 
a_i^{(w)}\exp\{\Delta_{i}^{(w)}\}, a_i^{(h)}\exp\{\Delta_{i}^{(h)}\})$
with the score being the multiplication of the objectness score and the conditional classification score. The final prediction results are the top-k (typically k = 300) candidate boxes sorted by the scores, after applying class specific Non-Maximum Suppression (NMS) with IoU threshold as 0.45.
We implement the approach on Caffe~\cite{jia2014caffe}.

\subsection{Experiment Results}
\subsubsection{PASCAL VOC}

\begin{table*}[t]
    \caption{Detection results on Pascal VOC 2007 test set, trained on VOC 07+12 trainval sets.
	Size represents the shorter edge of test image size. mAP\textsubscript{.5} stands for mean average precision at IoU 0.5. AP for each class is also reported.  }
    \small
    \setlength{\tabcolsep}{1pt}
	\begin{tabular}{lcc *{20}{c}}
		\toprule
		Method & Size & mAP\textsubscript{.5} & aero & bike & bird & boat & bottle & bus & car & cat & chair & cow & table & dog & horse & mbike & pers & plant & sheep & sofa & train & tv \\
		\midrule
		faster rcnn vgg\cite{Ren2015FasterRT} & 600 & 73.2 & 76.5 & 79.0 & 70.9 & 65.5 & 52.1 & 83.1 & 84.7 & 86.4 & 52.0 & 81.9 & 65.7 & 84.8 & 84.6 & 77.5 & 76.7 & 38.8 & 73.6 & 73.9 & 83.0 & 72.6 \\
		faster rcnn res\cite{He2016DeepRL} & 600 & 76.4 & 79.8 & 80.7 & 76.2 & 68.3 & 55.9 & 85.1 & 85.3 & 89.8 & 56.7 & 87.8 & 69.4 & 88.3 & 88.9 & 80.9 & 78.4 & 41.7 & 78.6 & 79.8 & 85.3 & 72.0 \\
		SSD512 \cite{Liu2016SSDSS} & 512 & 76.8 & 82.4 & 84.7 & 78.4 & 73.8 & 53.2 & 86.2 & 87.5 & 86.0 & 57.8 & 83.1 & 70.2 & 84.9 & 85.2 & 83.9 & 79.7 & 50.3 & 77.9 & 73.9 & 82.5 & 75.3 \\
		YOLOv2 \cite{Redmon2017YOLO9000BF} & 416 & 76.8 &-&-&-&-&-&-&-&-&-&-&-&-&-&-&-&-&-&-&-&- \\
		YOLOv2 \cite{Redmon2017YOLO9000BF} & 544 & 78.6 &-&-&-&-&-&-&-&-&-&-&-&-&-&-&-&-&-&-&-&- \\
		\midrule
		Baseline (identical) & 416 & 75.76 & 75.6 & 84.2 & 77.0 & 63.0 & 47.3 & 82.8 & 84.1 & 90.6 & 55.2 & 80.8 & 72.5 & 86.3 & 87.4 & 84.6 & 75.9 & 48.0 & 79.1 & 77.2 & 85.8 & 77.9 \\
		Baseline (uniform) & 416 & 76.32 & 75.9 & 83.7 & 75.1 & 64.1 & 50.5 & 84.3 & 83.9 & 91.4 & 57.7 & 81.8 & 73.7 & 88.6 & 88.0 & 83.8 & 77.1 & 47.6 & 77.0 & 78.4 & 88.1 & 75.8 \\
		Baseline (k-means) & 416 & 76.83 & 76.9 & 85.1 & 76.3 & 63.8 & 46.8 & 83.6 & 83.4 & 91.4 & 56.4 & 84.8 & 77.3 & 88.5 & 88.2 & 83.5 & 77.2 & 50.3 & 80.2 & 81.2 & 86.6 & 75.3 \\
		Baseline (k-means) & 544 & 79.45 & 77.5 & 87.2 & 80.1 & 66.5 & 56.1 & 85.3 & 86.2 & 89.7 & 63.0 & 88.6 & 76.5 & 88.0 & 91.0 & 87.9 & 81.9 & 53.8 & 84.9 & 79.5 & 86.1 & 79.5 \\
        \midrule
        Opt (identical)    & 416 & 78.01 & 77.8 & 86.6 & 78.3 & 67.6 & 50.6 & 85.1 & 85.1 & 91.6 & 59.1 & 82.3 & 78.0 & 88.5 & 90.2 & 86.2 & 79.0 & 53.1 & 81.4 & 81.9 & 89.5 & 76.3 \\
        Opt (uniform) & 416 & 77.95 & 78.2 & 87.3 & 75.3 & 67.2 & 52.9 & 86.3 & 85.3 & 90.5 & 56.2 & 84.1 & 76.9 & 89.7 & 89.5 & 85.9 & 78.9 & 50.2 & 79.2 & 82.1 & 87.1 & 76.9 \\
        Opt (k-means) & 416 & 77.99 & 76.3 & 87.4 & 77.6 & 66.6 & 52.0 & 85.3 & 85.0 & 91.5 & 57.5 & 83.6 & 77.1 & 88.6 & 90.6 & 84.7 & 78.4 & 50.2 & 82.7 & 80.3 & 87.1 & 76.6 \\
        Opt (k-means) & 544 & \textbf{80.69} & 75.8 & 88.3 & 79.4 & 66.8 & 56.9 & 88.5 & 87.9 & 89.6 & 62.4 & 88.8 & 75.4 & 89.0 & 90.9 & 88.7 & 83.2 & 51.1 & 84.7 & 73.2 & 86.6 & 80.3 \\
		\bottomrule
	\end{tabular}
	\label{tbl_voc}
\end{table*}

\begin{figure}[b]
\begin{center}
\includegraphics[width=0.88\linewidth]{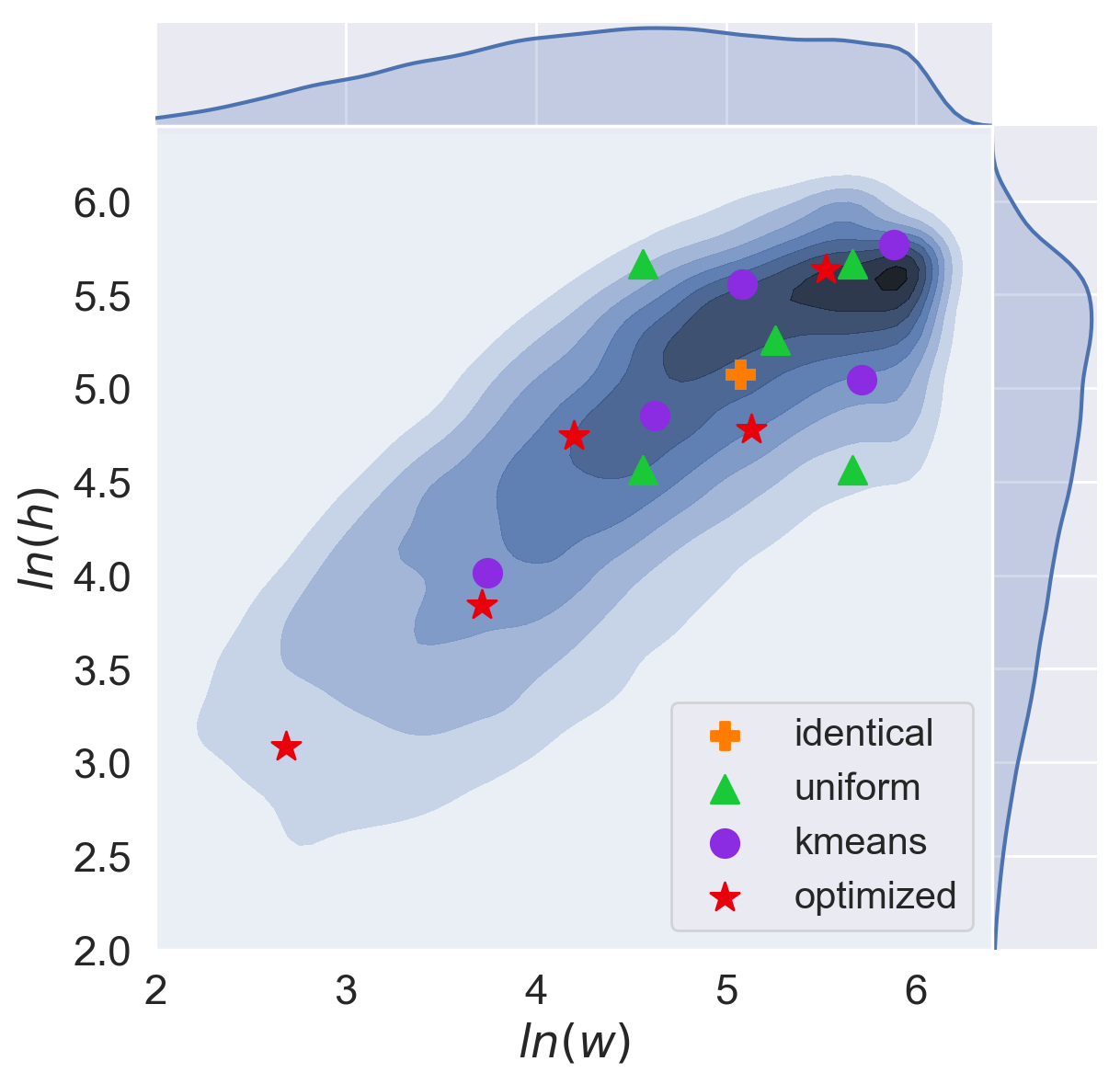}
\end{center}
\caption{Pascal VOC anchors and box distribution in log scale. The markers refer to the anchor shapes of different methods. Underlying the markers is the kernel density plot of the bounding box widths and heights with image resized to $416\times 416$. Darker color means higher density. Around are the marginal distributions of log(w) and log(h). }
\label{fig_voc_anc}
\end{figure}

The PASCAL VOC dataset\cite{Everingham2009ThePV}  contains box annotations over 20 object categories. We adopt the commonly used 07+12 train/test split, where the VOC 2007 \verb|trainval| (5k images) and VOC 2012 \verb|trainval| (11k images) are used as train set, and VOC 2007 \verb|test| (4952 images) is used as test set. The model is trained with 30,000 iterations of SGD (Momentum = 0.9) with mini-batches of 64 images evenly divided onto 4 GPUs. The learning rate is set to step-wise schedule: (0$\sim$100,1e-4), (100$\sim$15000,1e-3), (15000$\sim$27000,1e-4), (27000$\sim$30000, 1e-5). 
The training image height and width is 416 or 544 pixels, matching the test image size.

We investigate three different ways of the anchor shape initialization to study the robustness:
\begin{enumerate}[itemsep=-1pt]
    \item uniform: The anchor shapes are chosen uniformly, i.e. $[(3,3), (3,9), (9,9), (9,3), (6,6)] \times \text{\emph{stride}}$, with the stride being 32 here. 
    \item identical: All 5 anchors boxes are identical and initialized as $ (5,5) \times \text{\emph{stride}}$. 
    \item $k$-means: The values are borrowed from the official code of YOLOv2 
    which are the $k$-means centers of the ground-truth box shapes with IoU as the metric.  
\end{enumerate}

The results are shown in Table~\ref{tbl_voc}. We also list Faster R-CNN and SSD results for completeness. Note that we mainly show the effectiveness of the proposed approach, rather than target for the best accuracy. In the Baseline (*) rows, the anchor shapes are fixed as in traditional detector training, whereas in the Opt (*) rows, the anchor shapes are optimized with the proposed method. Anchor optimization consistently produces better results compared to the baselines. The re-implementations of YOLOv2 attain similar or better performances compared to those in the original paper (comparing Baseline (k-means) with YOLOv2). Our anchor optimization method boosts the performance by more than 1.2\% in terms of absolute mAP values. For example, with $k$-means initialization and 544 as the image size, the baseline achieves 79.45\% mAP, while our method improves the accuracy to 80.69\%.
Furthermore, different anchor shape initializations achieve similar accuracy. Within Opt (*), the accuracy difference between the best and the worst is only 0.06\% at size 416, suggesting that our method is very robust to different initial anchor configurations. Hence, the manual choice of appropriate initial anchor shapes becomes less critical. Note for the {\it{identical}} setting, although the anchor shapes are the same at the beginning, they can be optimized to take different values since they become responsible for different ground-truth boxes as training progresses. 

Figure~\ref{fig_voc_anc} illustrates the uniform anchors, the k-means anchors, the learned anchors (with $k$-means initialization), and the ground-truth box distribution in the $\log w$-$\log h$ plane. We observe that both the learned anchors and the k-means anchors align closely with the underlying ground truth box distribution, which intuitively explains why they achieve better performance. The learned anchors spread broader and are slightly smaller than the k-means anchors. The cause might be that small boxes are relatively harder to regress and the training pushes the anchors to focus more on small objects to lower the loss. This suggests that the anchor optimization process is more than merely clustering. It is also able to adapt the anchor shapes to data augmentation and network regression capability to improve the accuracy.

\subsubsection{MS COCO}

\begin{table}[tb]
\small
    \caption{Detection results on MS COCO \texttt{val}. Average Precisions (AP) at different IoU thresholds and different box scales (Small, Medium, Large at IoU 0.5) are reported. }
	\setlength{\tabcolsep}{3pt}
 	\centering
	\begin{tabular}{l|ccc|ccc}
		\toprule
		Method & AP\textsubscript{.5:.95} & AP\textsubscript{.5} & AP\textsubscript{.75} & AP\textsubscript{.5S} & AP\textsubscript{.5M} & AP\textsubscript{.5L} \\
		\midrule
		Baseline(uniform) & 21.90 & 42.06 & 20.57  & 2.27 & 29.04 & 57.10 \\
		Baseline(k-means) & 23.45 & 43.87 & 22.84  & 2.55 & 31.00 & 59.43 \\
        \midrule
        Opt (identical)    & 24.43 & 45.07 & 24.05  & 3.03 & 32.24 & 60.61 \\
        Opt (uniform) & 24.55 & 45.33 & 24.04  & 3.11 & 32.52 & 60.70 \\
        Opt (k-means) & 24.47 & 45.07 & 23.74  & 3.06 & 32.65 & 60.09 \\
		\bottomrule
	\end{tabular}
	\label{tbl_coco}
	\vspace{-10pt}
\end{table}

\begin{table*}[t]
    \caption{Detection results from the evaluation server on MS COCO \texttt{test-dev}. AP means average precision, AR means average recall. AP\textsubscript{.5:.95} is the mean of AP at IoU 0.5:0.05:0.95. Subscript S,M \& L correspond to small, median \& large bounding boxes respectively. }
	\setlength{\tabcolsep}{6.1pt}
 	\centering
	\begin{tabular}{l|ccc|ccc|ccc|ccc}
		\toprule
		Method & AP\textsubscript{.5:.95} & AP\textsubscript{.5} & AP\textsubscript{.75} &
		AP\textsubscript S & AP\textsubscript M & AP\textsubscript L &
		AR\textsubscript 1 & AR\textsubscript{10} & AR\textsubscript{100} &
		AR\textsubscript{S} & AR\textsubscript{M} & AR\textsubscript{L} \\
		\midrule
		Faster RCNN vgg\cite{Ren2015FasterRT} & 21.9 & 42.7 & -  & - & - & -  & - & - & -  & - & - & - \\
		Faster RCNN \cite{coco_website} & 24.2 & 45.3 & 23.5 & 7.7 & 26.4 & 37.1 & 23.8 & 34.0 & 34.6 & 12.0 & 38.5 & 54.4 \\
		SSD512 \cite{Liu2016SSDSS} & 26.8 & 46.5 & 27.8  & 9.0 & 28.9 & 41.9  & 24.8 & 37.5 & 39.8  & 14.0 & 43.5 & 59.0 \\
		YOLOv2 \cite{Redmon2017YOLO9000BF} & 21.6 & 44.0 & 19.2  & 5.0 & 22.4 & 35.5  & 20.7 & 31.6 & 33.3  & 9.8 & 36.5 & 54.4 \\
		\midrule
		Baseline (uniform) & 22.4 & 42.5 & 21.4  & 4.4 & 21.5 & 38.8  & 21.5 & 31.2 & 32.1  & 7.3 & 32.9 & 57.4  \\
		Baseline (k-means) & 24.0 & 44.9 & 23.3  & 4.4 & 24.6 & 40.9  & 22.4 & 33.0 & 34.1  & 7.6 & 37.2 & 58.4  \\
        \midrule
        Opt (identical) & 25.0 & 45.8 & 24.5  & 5.7 & 26.5 & 40.4  & 23.3 & 34.4 & 35.6  & 9.5 & 39.4 & 58.8  \\
        Opt (uniform) & 25.0 & 45.8 & 24.3  & 5.9 & 26.1 & 40.8  & 23.3 & 34.4 & 35.6  & 9.5 & 39.0 & 59.1  \\
        Opt (k-means) & 25.0 & 45.9 & 24.7  & 5.7 & 26.6 & 40.8  & 23.3 & 34.4 & 35.6  & 9.5 & 39.6 & 58.8  \\
		\bottomrule
	\end{tabular}
	\label{tbl_coco_test}
	\vspace{-10pt}
\end{table*}

We adopt the widely used COCO \cite{Lin2014MicrosoftCC} 2017 Detection Challenge train/val splits, where the train set has 115K images, the \verb|val| set has 5K images, and the \verb|test-dev| set has about 20k images whose box annotations are not publicly available. The dataset contains 80 object categories.

We use similar training configurations as the VOC experiments. Mini-batch size is 64 and evenly split onto 4 GPUs. Momentum of SGD is set to 0.9. Since the COCO dataset has substantially more images than VOC, we increase the number of iterations to 100,000, and set the learning rate schedule to (0$\sim$1000,1e-4), (1000$\sim$80000,1e-3), (80000$\sim$90000,1e-4), (90000$\sim$100000, 1e-5). The train and test image sizes are both set to 544 in all experiments. The bounding box annotations of the \verb|test-dev| is not exposed. We upload our detection results to the official COCO evaluation server 
to retrieve the scores.

\begin{figure}[tb]
\begin{center}
\includegraphics[width=0.89\linewidth]{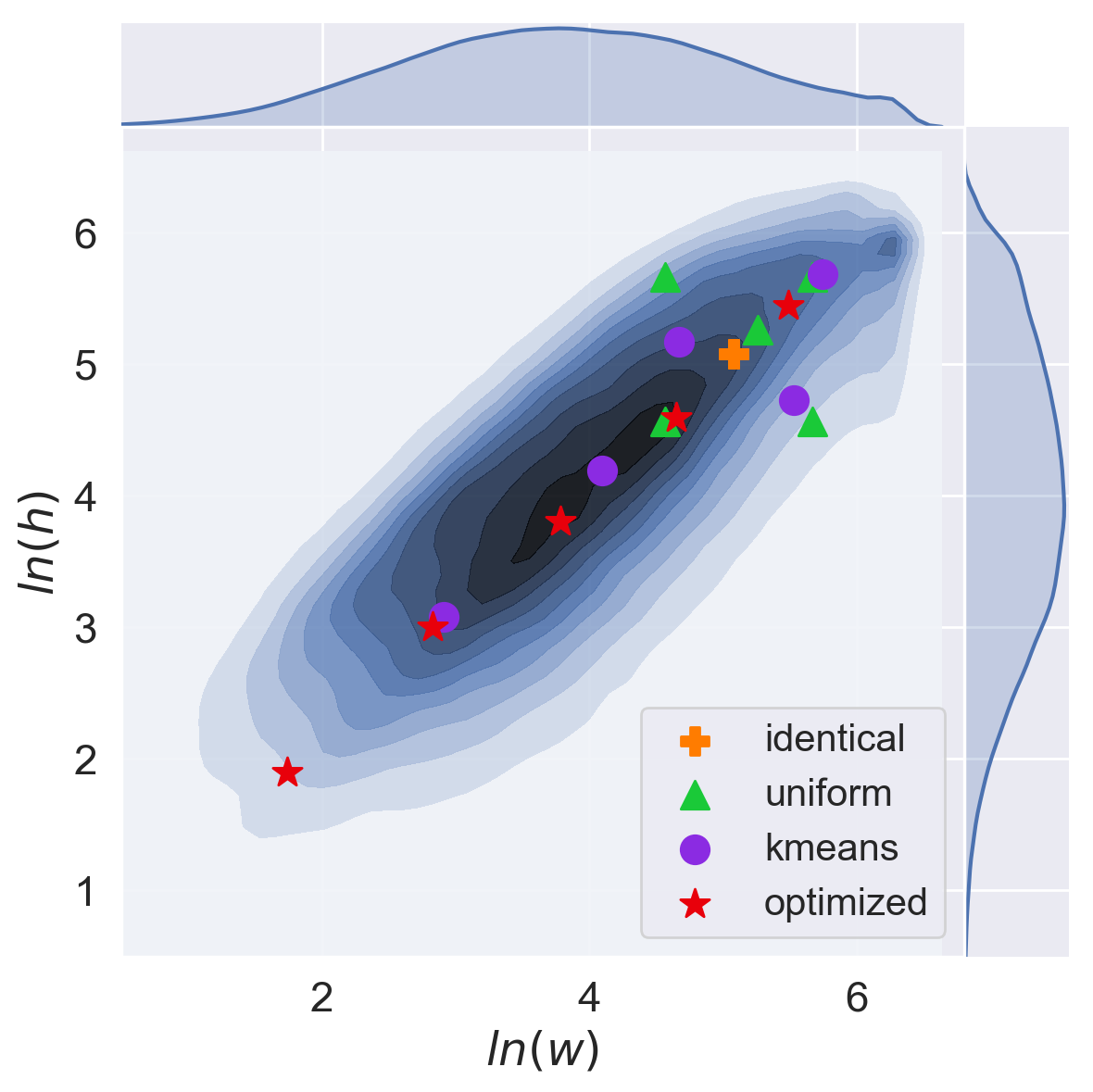}
\end{center}
\caption{COCO anchors and box distribution in log scale. Underlying the markers is the kernel density of the ground-truth bounding box widths and heights (images resized to 544x544). Around the figure are the marginal distributions of $\log(w)$ and $\log(h)$.}
\label{fig_coco_anc}
\vspace{-10pt}
\end{figure}

The results on the \verb|val| set are shown in Table~\ref{tbl_coco}, and the results on the \verb|test-dev| set are in Table~\ref{tbl_coco_test}. 
In the tables, AP\textsubscript{.5:.95} denotes the average of mAP evaluated at IoU threshold evenly distributed between 0.5 and 0.95; AR denotes the average recall rate. Compared to the original YOLOv2 results, our reimplementation achieves a higher accuracy with the AP\textsubscript{.5:.95} increased from 21.6\% to 24.0\% and AP\textsubscript{.5} from 44.0\% to 44.9\%. 
When equipped with the proposed anchor optimization method, the accuracy is further boosted by 1\%, with AP\textsubscript{.5:.95} to 25.0\% and AP\textsubscript{.5} to 45.9\%. Similar trends are observed from the \verb|val| split results. This strongly demonstrates the superiority of the anchor optimization to achieve higher accuracy. 
Meanwhile, the baseline approach without anchor optimization is quite sensitive to the anchor shapes. On \verb|val|, $k$-means initialization achieves 23.45\%, but the uniform initialization achieves only 21.90\% with a 1.55 points difference. On the contrary, our optimization approach is more robust and the difference between the highest (24.55) and the lowest (24.43) is only 0.12 on \verb|val|. On \verb|test-dev|, different initialization methods achieve the same mAP\textsubscript{.5:.95} (25.0), which further verifies the robustness towards different initialization methods.

The learned anchors with different initializations are shown in Table~\ref{tbl_coco_anchor}. We can observe that the anchor shapes are quite similar though their initial values are different. 
Figure~\ref{fig_coco_anc} shows the learned anchor shapes against the uniform and the k-means anchors. 
The learned anchors nicely cover the ground-truth bounding box distribution. They also tend to be slightly smaller than the original k-means values, which could help the small object detection since the large object is relatively easy to detect. This can also be verified from numbers in Table~\ref{tbl_coco_test}. Take, for instance, the $k$-means initialization, the gain from small (from 4.4\% to 5.7\%) and medium objects (from 24.6\% to 26.6\%) is high while it slightly sacrifices the accuracy for large objects (from 40.9\% to 40.8\%).

\begin{table}[t]
    \centering
    \caption{Learned anchors from different initializations on COCO with image size as $544$. } 	\label{tbl_coco_anchor}
 	\small
 	\setlength{\tabcolsep}{2.43pt}
	\begin{tabular}{lccccc}
		\toprule
		Init & $s_1^{(w)}, s_1^{(h)}$ & $s_2^{(w)}, s_2^{(h)}$ & $s_3^{(w)}, s_3^{(h)}$ & $s_4^{(w)}, s_4^{(h)}$ & $s_5^{(w)}, s_5^{(h)}$ \\
		\midrule
		identical & 5.8,    6.7 &   17.4,   20.1 &   44.8,   45.8 &  108,   99.2 &     241,  237 \\
        uniform   & 5.8,    6.8 &   17.4,   20.5 &   44.8,   45.8 &  106,  101 &     245,  237 \\
        k-means   & 5.7,    6.7 &   16.9,   20.1 &   43.8,   44.8 &  104,   98.9 &     241,  230 \\
		\bottomrule
	\end{tabular}
	\vspace{-15pt}
\end{table}

\subsubsection{Brainwash}
Brainwash is a head detection dataset introduced in \cite{Stewart2016EndtoEndPD} with about 10k training images, about 500 validation images and 484 testing images. The images are captured from a surveillance camera in a cafe shop
. We train the detection model for 10,000 steps, with learning rate schedule (0$\sim$100,1e-4), (100$\sim$5,000,1e-3), (5,000$\sim$9,000,1e-4), (9,000$\sim$10,000, 1e-5). No random scaling augmentation is used since the camera is still, while other kinds of data augmentation remain unchanged. The image crop size during training is 320, and the test image size is 640. We still choose to employ 5 anchor shapes. No classification loss is applied since there is only one category (head). 

We report AP\textsubscript{.5} as the performance criterion in Table~\ref{tbl_brain}. The baseline result with the anchor shapes from COCO is also presented. The $k$-means anchors are computed similarly to those in YOLOv2. Since the head bounding boxes are much smaller than those of COCO, we find that, with the COCO anchor shape settings, only one anchor shape (out of 5) will be activated throughout the training. In this case, the neural network will also need to predict large deviations for $w$ and $h$ to fit all the ground-truth boxes, which can be sub-optimal.
This means it can be sub-optimal to directly use the anchor shapes from other domains. 
Comparatively, the proposed anchor learning method can adapt the anchor shapes quickly to this dataset. From the results, we observe Opt (*) consistently outperform the baselines by large margins, demonstrating the effectiveness of the proposed method.
Even with the $k$-means as the initialized anchor shapes, our approach can further improve the accuracy by 1.2 points (from 78.98\% to 80.18\%).

\begin{table}[t]
    \caption{Detection results on Brainwash dataset. Test image size is 640. AP\textsubscript{.5} is the average precision with IoU threshold 0.5. }
 	\centering
	\begin{tabular}{lcc}
		\toprule
		Method & Size & AP\textsubscript{.5} \\
		\midrule
		Baseline (coco) & 640 & 77.96 \\
		Baseline (uniform) & 640 & 78.03 \\
		Baseline (k-means) & 640 & 78.98 \\
        \midrule
        Opt (identical) & 640 & 79.85 \\
        Opt (uniform) & 640 & 79.86 \\
        Opt (k-means) & 640 & 80.18 \\
		\bottomrule
	\end{tabular}
	\label{tbl_brain}
	\vspace{-8pt}
\end{table}


\subsection{Ablation studies}

\textbf{Training techniques.} The first experiment is to evaluate the effectiveness of the three training techniques: 
soft assignment warm-up (Soft),
online clustering warm-up (Clust),
and batch normalization without shifting (BN).
Table~\ref{tab:ablation_tricks} shows the experimental 
results on the Pascal VOC 2007 and Brainwash datasets. 
Without any of these three techniques, our optimization approach surpasses the baseline by 0.23 point on   
VOC 2007 and by 0.86 point
on Brainwash respectively. 
With the three techniques to facilitate 
training, the gain gradually increases to 1.16
and 1.83 points respectively. We want to note that the BN trick alone does not bring any noticeable change to the performance of the Baseline method ($<0.1\%$ mAP), suggesting that it has to be combined with anchor optimization to be effective.

\begin{figure}
    \centering
    \includegraphics[width=0.8\linewidth]{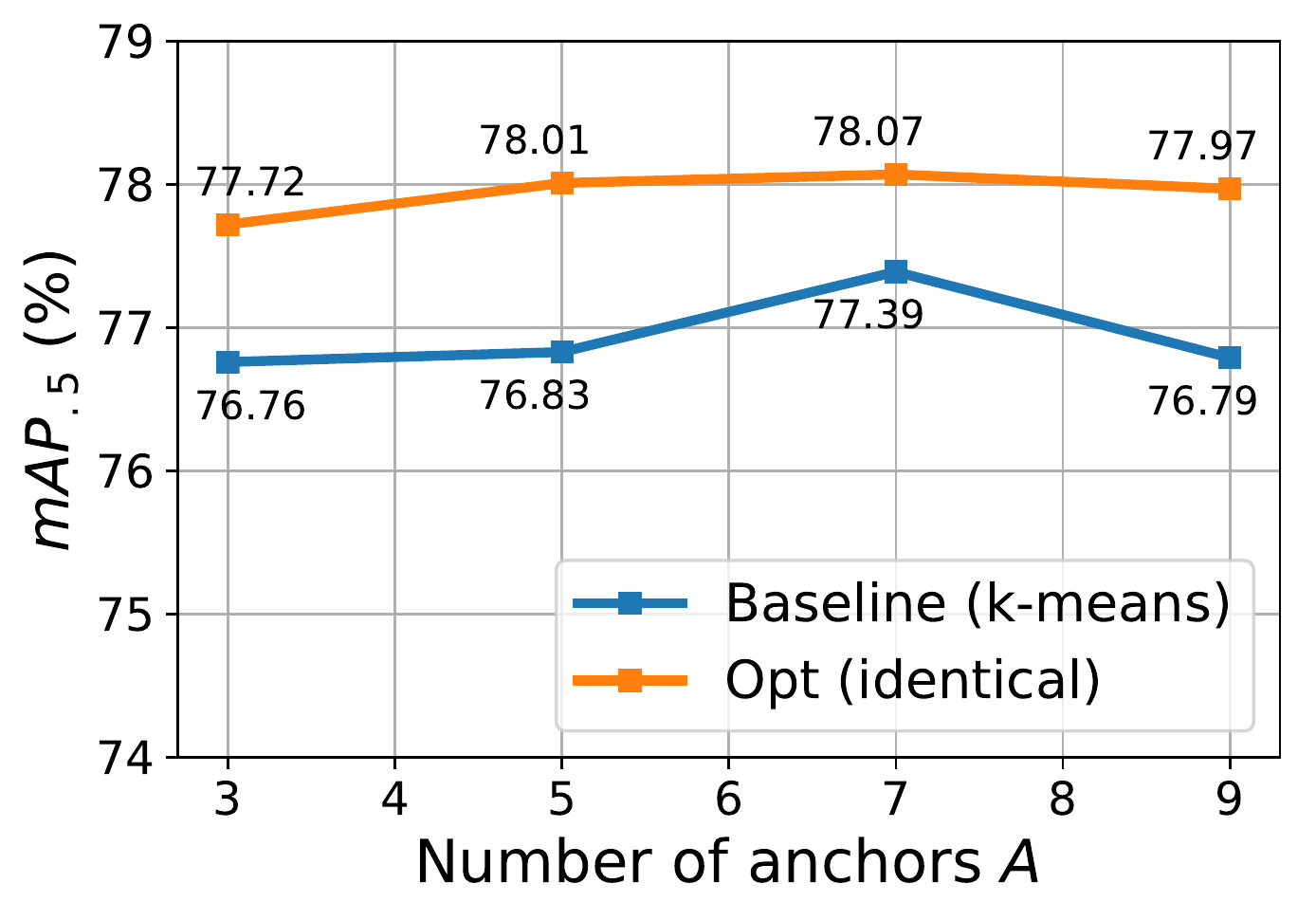}
    \caption{VOC07 test performance, varying number of anchors $A$. }
    \label{fig_varya}
\end{figure}

\textbf{Number of anchors.} The second experiment is to show the robustness of the 
improvement across different numbers of anchors, 
illustrated in Fig.~\ref{fig_varya}. 
With different numbers of anchor shapes, the accuracy with
optimized anchor shapes consistently outperforms the 
baseline by around $0.5\sim 1.2\%$. The performance drop of 9 relative to 7 anchors could be caused by insufficient training, since the number of examples each anchor shape gets is reduced. Nevertheless, the choice of number of anchors does not affect the improvement of the proposed method over the baseline.

\begin{table}[b]
    \centering
    \small
    \caption{Ablation study on the training techniques on the Pascal VOC 2007 and Brainwash datasets in terms of mAP\textsubscript{.5} (\%) .}
    \begin{tabular}{l|cc}
        \toprule
        Method    & Pascal VOC 2007&  Brainwash \\
        \midrule
        Baseline            & 76.83 (+0.00)& 78.03 (+0.00) \\
        Opt                 & 77.06 (+0.23)& 78.89 (+0.86) \\
        Opt+Soft            & 77.38 (+0.55)& 79.12 (+1.09) \\
        Opt+Soft+Clust      & 77.56 (+0.73)& 79.42 (+1.39) \\
        Opt+Soft+Clust+BN   & \textbf{77.99 (+1.16)}& \textbf{79.86 (+1.83)} \\
        \bottomrule
    \end{tabular}
    \label{tab:ablation_tricks}
    \vspace{-5pt}
\end{table}

\begin{table}[t]
\centering
    \caption{Faster-RCNN. mAP@.5 for VOC, mAP\textsubscript{.5:.95} for COCO. }
	\setlength{\tabcolsep}{3pt}
	\begin{tabular}{lccc}
		\toprule
		Model & Dataset & Baseline & Ours \\
		\midrule 
		ResNet50 & VOC 07    & 73.03 & 73.59 (+0.56) \\
		ResNet50 & VOC 07+12 & 78.52 & 79.02 (+0.50) \\
		ResNet50 FPN & VOC 07+12 & 80.94 & 81.46 (+0.52) \\
		ResNet50 FPN & COCO 2017 & 36.78 & 37.09 (+0.31) \\
		\bottomrule
	\end{tabular}
	\label{tbl_frcnn}
	\vspace{-10pt}
\end{table}

\textbf{Faster RCNN.} Finally, we apply our anchor box optimization approach (without BN or soft assignment trick) to the Faster-RCNN framework with ResNet50 to evaluate the effectiveness on two-stage detectors. Since the batch size is small, BN tends to be unstable thus is not adopted. Anchor boxes are used in the Region Proposal Network (RPN) to generates candidate bounding boxes, which are refined and classified in the second stage. We observe consistent improvements across different datasets and architectures (w/ or w/o FPN) in Table~\ref{tbl_frcnn} results.
Meanwhile, the recall of RPN on COCO 2017 is increased from 41.77\% to 42.60\% (+0.83\%) at top 50; 47.77\% to 48.28\% (+0.51\%) at top 100.
The final improvement is not as significant as that of the one-stage detector and the reason might be that the second stage regression reduces the negative effect of improper anchor sizes in the first stage. However, it is worth noting that the gain comes from negligible extra training cost and no extra inference cost. In fact, for Faster-RCNN R50 FPN on COCO, the training time per iteration is 0.8154s compared to the original 0.8027s on 4 GPUs with pytorch.


\section{Conclusion}\label{sec:conclusion}

In this paper, we have introduced an anchor optimization method which can be employed in most existing anchor-based object detection frameworks to automatically learn the anchor shapes during training. The learned anchors are better suited for the specific data and network structure and can produce better accuracy. We demonstrated the effectiveness of our method mainly on the popular one-stage detector YOLOv2. Extensive experiments on Pascal VOC, COCO and Brainwash datasets show superior detection accuracy of our method over the baseline. We also show that the anchor optimization method is robust to initialization configurations, and hence the careful handcrafting of anchor shapes is greatly alleviated for good performance.

{\small
\bibliographystyle{ieee}
\bibliography{egbib}
}

\end{document}